\documentclass[letterpaper]{article} % DO NOT CHANGE THIS
\usepackage{aaai2026}  % DO NOT CHANGE THIS
\usepackage{times}  % DO NOT CHANGE THIS
\usepackage{helvet}  % DO NOT CHANGE THIS
\usepackage{courier}  % DO NOT CHANGE THIS
\usepackage[hyphens]{url}  % DO NOT CHANGE THIS
\usepackage{graphicx} % DO NOT CHANGE THIS
\usepackage{natbib}  % DO NOT CHANGE THIS AND DO NOT ADD ANY OPTIONS TO IT
\usepackage{caption} % DO NOT CHANGE THIS AND DO NOT ADD ANY OPTIONS TO IT
\usepackage{algorithm}
\usepackage{algorithmic}

\usepackage{booktabs}
\usepackage{amsmath}
\usepackage{amssymb}
\usepackage{xcolor}

\usepackage{newfloat}
\usepackage{listings}
\DeclareCaptionStyle{ruled}{labelfont=normalfont,labelsep=colon,strut=off} % DO NOT CHANGE THIS
\floatstyle{ruled}
\newfloat{listing}{tb}{lst}{}
\floatname{listing}{Listing}
\title{Beyond Accuracy: A Cognitive Load Framework for Mapping the Capability Boundaries of Tool-use Agents}
\author{
    Qihao Wang\textsuperscript{\rm 1,2},
    Yue Hu\textsuperscript{\rm 1,2*},
    Mingzhe Lu\textsuperscript{\rm 1,2},
    Jiayue Wu\textsuperscript{\rm 1,2},
    Yanbing Liu\textsuperscript{\rm 1,2*},
    Yuanmin Tang\textsuperscript{\rm 1,2*}
}
\affiliations{
\textsuperscript{\rm 1}Institute of Information Engineering, Chinese Academy of Sciences \\
\textsuperscript{\rm 2}School of Cyber Security, University of Chinese Academy of Sciences \\
}

\begin{document}

\maketitle

\begin{abstract}
The ability of Large Language Models (LLMs) to use external tools unlocks powerful real-world interactions, making rigorous evaluation essential. However, current benchmarks primarily report final accuracy, revealing what models can do but obscuring the cognitive bottlenecks that define their true capability boundaries. To move from simple performance scoring to a diagnostic tool, we introduce a framework grounded in Cognitive Load Theory. Our framework deconstructs task complexity into two quantifiable components: Intrinsic Load, the inherent structural complexity of the solution path, formalized with a novel Tool Interaction Graph; and Extraneous Load, the difficulty arising from ambiguous task presentation. To enable controlled experiments, we construct ToolLoad-Bench, the first benchmark with parametrically adjustable cognitive load. Our evaluation reveals distinct performance cliffs as cognitive load increases, allowing us to precisely map each model's capability boundary. We validate that our framework's predictions are highly calibrated with empirical results, establishing a principled methodology for understanding an agent's limits and a practical foundation for building more efficient systems.
\end{abstract}

\section{Introduction}

The paradigm of Large Language Models (LLMs) is rapidly evolving from passive text generators into capable autonomous agents that can interact with the world and solve complex problems \cite{yaoreact, hugginggpt, wang2024survey, mialon2023augmented}. A cornerstone of this transformation is the ability to use external tools, APIs, databases, and other software to overcome the limitations of their parametric knowledge \cite{schick2023toolformer, mialon2023augmented, qin2024tool, su2025openthinkimg}. This tool-use capability is the engine driving progress in agentic AI, enabling models to tackle multi-step, real-world tasks that were previously intractable \cite{chen2025tool, shi2025tool}.

This rapid progress has spurred the development of sophisticated benchmarks designed to measure and rank the tool-use proficiency of different models. Influential testbeds like \citet{api_bank} and \citet{toolbench} provide comprehensive evaluations with thousands of APIs and complex tasks. The Berkeley Function Calling Leaderboard \cite{patil2025bfcl} further standardizes evaluation, while newer benchmarks explore more realistic conversational \cite{tooltalk, du2024anytool} and stateful \cite{shi2024learning} scenarios. While these evaluations are invaluable for tracking overall progress, they typically culminate in a single, final accuracy score. This black-box evaluation paradigm reveals what a model can achieve but obscures when it fails. They treat task difficulty as a singular, unanalyzed variable, lacking a granular framework to diagnose specific failure modes related to task complexity. Consequently, there is a lack of clear understanding of the specific operational limits the capability boundaries of these agents \cite{qu2025tool, chen2025toleap}.

To move beyond simple performance scoring towards a more diagnostic form of evaluation, we propose a new lens inspired by Cognitive Load Theory (CLT) from psychology \cite{sweller1988cognitive, plass2010cognitive}. Our approach deconstructs task complexity into quantifiable components, allowing us to systematically probe an agent's capabilities under controlled conditions. This enables us to pinpoint the specific bottlenecks that limit an agent's performance, a departure from prior work that has used CLT primarily as a metaphor for computational cost or to reduce the cognitive burden on human users \cite{yang2025sparse, guidroz2025llm}.

Our framework operationalizes this by modeling an agent's performance as a function of cognitive load. This allows us to characterize each agent's capability not as a single point of accuracy, but through a more descriptive cognitive profile defined by two key parameters: its Baseline Capability, reflecting its intrinsic proficiency on low-complexity tasks, and its Load Sensitivity, which measures how gracefully its performance degrades as task complexity increases. Crucially, we validate the soundness of our theoretical model through rigorous statistical goodness-of-fit tests, confirming that its predictions are highly calibrated with empirical observations.

Our primary contributions are as follows:

\begin{enumerate}
    \item \textbf{A Formal Cognitive Load Framework:} We propose and formalize a novel evaluation framework, grounded in Cognitive Load Theory, that deconstructs task difficulty into quantifiable components: Intrinsic Load, derived from the inherent structural complexity of the task, and Extraneous Load, arising from the ambiguity of its presentation.
    
    \item \textbf{A Parametrically-Controlled Benchmark:} We construct and release \textbf{ToolLoad-Bench}, the first benchmark designed for controlled experimentation, with instances that allow for the parametric adjustment of cognitive load to systematically probe model limits.
    
    \item \textbf{Empirical Mapping of Capability Boundaries}: We conduct an extensive evaluation of leading models, using our framework to map their distinct capability boundaries. This analysis reveals performance cliffs and characterizes each agent with a unique cognitive profile based on its baseline capability and load sensitivity.
    
    \item \textbf{A Validated Evaluation Methodology:} We validate our framework's predictive power, showing through statistical testing that it is well-calibrated with empirical results. This establishes a principled and diagnostic methodology for the future assessment of tool-use agents.
\end{enumerate}
\section{Related Work}

\subsection{Enhancing Tool-Use Capabilities}
Research in tool-augmented LLMs has rapidly advanced, focusing heavily on improving model proficiency. Foundational work demonstrated that models could learn to use tools through self-supervised learning \cite{schick2023toolformer}. This was followed by a wave of instruction tuning, using curated datasets to teach models specific tool-use formats and behaviors \cite{toolalpaca, gorilla, lin2024hammer, shen2024small}. To overcome the bottleneck of manual data creation, recent efforts have focused on automated data generation pipelines that simulate agentic interactions to produce complex, multi-turn training data \cite{prabhakar2025apigen, shi2025taskcraft, yin2025magnet, zhang2025nemotron}. Concurrently, reinforcement learning (RL) has emerged as a powerful technique to refine tool-use policies, optimizing for reward signals related to successful task completion and reasoning \cite{qian2025toolrl, dong2025tool, dong2025agentic, wang2025otc, feng2025retool}. Our work complements these advancements by providing a more nuanced evaluation framework to measure the true capabilities of the agents they produce.

\subsection{Cognitive Load Theory in LLM Research}
Originating in educational psychology, Cognitive Load Theory (CLT) posits that learning is impeded when a task's demands exceed the finite capacity of working memory \cite{sweller1988cognitive}. This theory has recently been adapted to LLM research in two main ways. First, it serves as a metaphor for computational cost, inspiring systems that dynamically manage model resources to improve efficiency, akin to reducing "cognitive" strain \cite{yang2025sparse, xiao2025streaming}. Second, in its traditional sense, CLT guides the design of LLM applications that reduce the cognitive burden on human users, for example, by simplifying text or generating helpful explanations \cite{guidroz2025llm, sirbu2025explanation}. Our work charts a new course by being the first to apply CLT not to system efficiency or human-computer interaction, but as a formal, quantitative framework to deconstruct task complexity and evaluate the cognitive limits of the AI agents themselves.

\section{Methodology}
\label{sec:framework}

\begin{figure*}[t]
    \centering
    \includegraphics[width=0.9\textwidth]{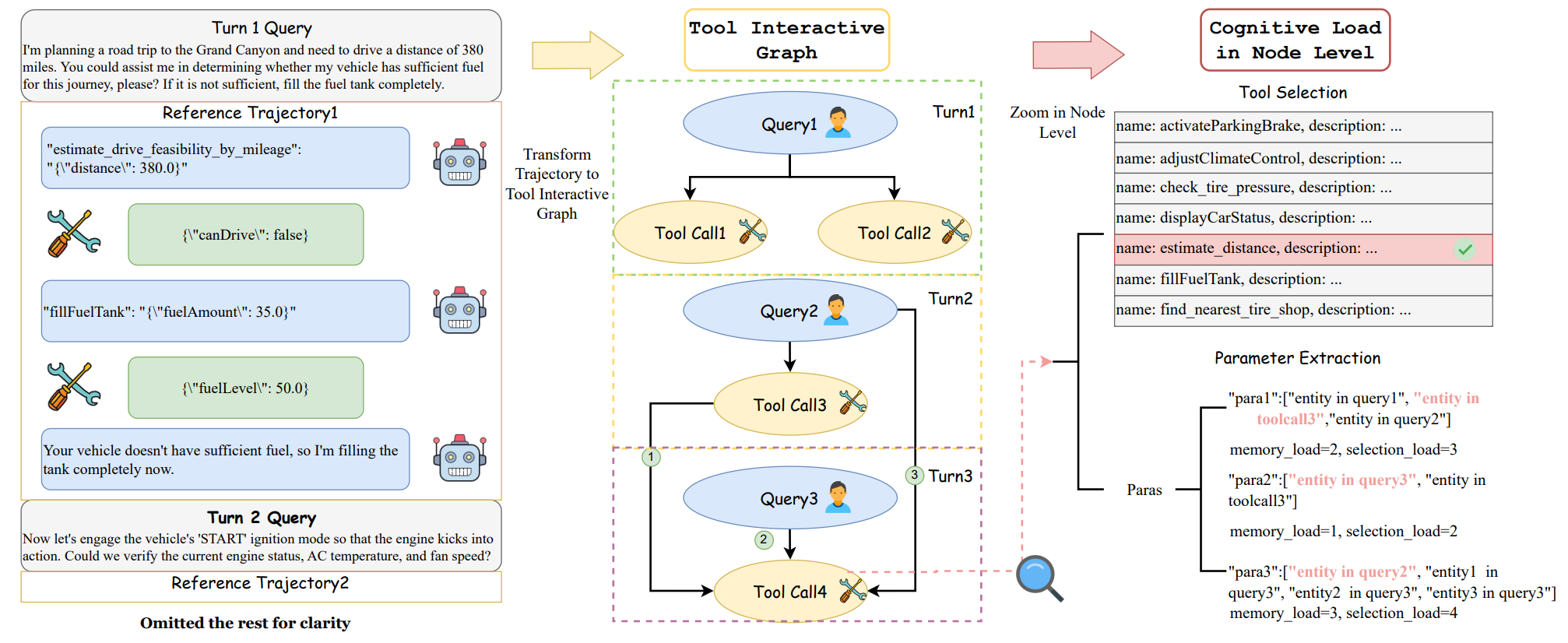}
    \caption{
An illustration of our Cognitive Load Framework. A multi-turn tool-use task, defined by a sequence of user queries and a set of tools , is mapped to a TIG. The TIG represents the ground-truth solution. Zooming in node level, It shows selection load in parameter extraction and extraneous cognitive Load in tool selection.}
    \label{fig:tig_framework}
\end{figure*}

To quantify task difficulty, we introduce a framework grounded in a probabilistic view of task success. We begin from formally defining the problem setting and our notation.

\subsection{Preliminary}
Our research focus on the problem of multi-turn tool use. A task instance is defined by a tuple $(Q, T)$, where:
\begin{itemize}
    \item $Q = (q_1, q_2, \dots, q_m)$ is an ordered sequence of user queries.
    \item $T = \{tool_1, tool_2, \dots, tool_k\}$ is the set of available API tools that an agent can use to finish the given task.
\end{itemize}
The tool-use agents' objective is to generate the correct sequence of tool calls from $T$ with  correct parameters.

\subsection{The Tool Interaction Graph}
To formally represent the internal complexity of a task's solution, we introduce the \textbf{Tool Interaction Graph (TIG)}. For a given instance $(Q, T)$, its TIG, denoted as $G=(V, E)$, is a directed acyclic graph (DAG) that models the ground-truth solution.

\begin{itemize}
    \item \textbf{Nodes ($V$):} The set of nodes consists of user query nodes $\{v_{q_1}, \dots, v_{q_m}\}$ and a set of function call nodes $\{v_{f_1}, \dots, v_{f_n}\}$. Each $v_{f_i}$ corresponds to a specific tool invocation required to solve the task.
    \item \textbf{Edges ($E$):} A directed edge $(v_i, v_j) \in E$ represents a dependency, which can be one of two types:
        \begin{itemize}
            \item \textbf{Data Dependency:} The invocation of function $v_j$ requires specific data produced by the operation at node $v_i$.
            \item \textbf{Execution Dependency:} A procedural constraint where task logic requires $v_i$ to be executed before $v_j$, even if no data is passed.
        \end{itemize}
\end{itemize}
The TIG provides a complete, formal scaffold of the task's inherent structural complexity, upon which we build our theory of cognitive load.

\subsection{A Primer on Cognitive Load Theory}
Cognitive Load Theory (CLT), originating from educational psychology, posits that human working memory has a limited capacity \cite{sweller1988cognitive}. Effective learning and problem-solving are hindered when the total cognitive demand of a task exceeds this capacity. CLT traditionally deconstructs this total load into two primary components relevant to our work:
\begin{itemize}
    \item \textbf{Intrinsic Cognitive Load:} The inherent, irreducible complexity of the subject matter itself, determined by the number of interacting elements that must be processed simultaneously. In our context, this corresponds to the structural complexity of the task's solution path, which we formalize using the Tool Interaction Graph (TIG).
    \item \textbf{Extraneous Cognitive Load:} The cognitive burden imposed by the way information is presented. This load is not essential to the task itself but arises from suboptimal design, such as confusing instructions or distracting information. We map this directly to challenges like ambiguous user queries and the presence of irrelevant distractor tools.
\end{itemize}
Our framework adapts this established psychological model to the domain of tool-augmented LLMs. We treat the model's computational context and reasoning capacity as analogous to human working memory. This allows us to move beyond a monolithic view of task difficulty and instead provide a principled, quantitative decomposition. By measuring intrinsic ($CL_I$) and extraneous ($CL_E$) load, we can precisely diagnose the specific bottlenecks that limit an agent's performance.

\subsection{Theoretical Postulates and Derivation of Additive Load}
Our framework relies on two fundamental postulates that connect cognitive load to the observable accuracy.

\paragraph{Postulate 1: The Load-Success Relationship.} We posit that the probability of successfully executing any single cognitive operation is an exponential function of its associated cognitive load, $CL$.
\begin{equation}
    P_{succ}(\text{op}) = \exp(-(k \cdot CL + b))
    \label{eq:load_success_relation}
\end{equation}
Here, $k > 0$ and $b \ge 0$  are  model-specific sensitivity parameter, representing model's capability to solve these tasks.

\paragraph{Postulate 2: Probabilistic Independence in the TIG.} The probability of successfully executing the entire plan is the product of the probabilities of successfully executing each constituent function call node $v_f$.
\begin{equation}
    P_{succ}(G) = \prod_{v_f \in V \setminus V_Q} P_{succ}(v_f)
    \label{eq:multiplicative_prob_dag}
\end{equation}
where $V_Q$ is the set of all query nodes.

\paragraph{Derivation.} From these postulates, we can derive the additive nature of cognitive load. By substituting the load-success relationship Equation \eqref{eq:load_success_relation} into the probabilistic independence model of Equation \eqref{eq:multiplicative_prob_dag}, the product of exponential terms becomes the sum of their exponents. This leads directly to our central proposition:
\begin{equation}
    CL_{Total} = \sum_{v_f \in V \setminus V_Q} CL(v_f)
    \label{eq:additive_load_principle}
\end{equation}
\textbf{Proposition 1.} \label{prop:additive_load} \textit{Given our postulates, the total cognitive load of a task is the sum of the cognitive loads of its constituent function call nodes.}

\subsection{Decomposing and Quantifying Cognitive Load}
This additive principle allows us to decompose the total load into its constituent sources: intrinsic load from the task's structure and extraneous load from its presentation.
\begin{equation}
    CL_{Total} = CL_I + CL_E
    \label{eq:total_load_decomposition}
\end{equation}

\subsubsection{Intrinsic Cognitive Load ($CL_I$)}
The intrinsic load is inherent to the TIG structure. Following Proposition \ref{prop:additive_load}, we define $CL_I$ as the sum of loads from all function nodes, which in turn is the sum of the loads of their dependency edges, $CL(e)$:
\begin{equation}
    CL_I(G) = \sum_{v_f \in V \setminus V_Q} \sum_{e=(v_i, v_f) \in E} CL(e)
    \label{eq:cli_sum_of_edges}
\end{equation}
The load of a single dependency edge, $CL(e)$, is determined by its difficulty, which we model as a weight, $w(e) = CL(e)$, composed of two factors:
\begin{itemize}
    \item \textbf{Memory Load (Attentional Distance):} The effort to recall information. We model this with $\delta(v_i, v_j)$, the number of conversational turns (user queries and tool calls) between the operation at $v_i$ and its use at $v_j$.
    \item \textbf{Selection Load (Interference):} The effort to select correct information. We model this with $I(v_i, v_j)$, the number of other available but incorrect entities of the same semantic type (e.g., other user IDs) in the context. For pure execution dependencies, this is zero.
\end{itemize}
These combine into a single edge weight:
\begin{equation}
    w(e) = \delta(v_i, v_j) \cdot (1 + \lambda \cdot I(v_i, v_j))
    \label{eq:edge_weight_detailed}
\end{equation}
where $\lambda$ is a balancing hyperparameter. Our final formulation for intrinsic load is:
\begin{equation}
    CL_I(G) = \sum_{v_f \in V \setminus V_Q} \sum_{e=(v_i, v_f) \in E} w(e)
    \label{eq:cli_final_formal}
\end{equation}

\subsubsection{Extraneous Cognitive Load ($CL_E$)}
Extraneous load arises from how the task is presented. It is independent of the TIG's structure and is primarily incurred when parsing user queries. We define the total extraneous load as the sum of the loads from each individual query in the task sequence $Q$:
\begin{equation}
    CL_E(Q, T) = \sum_{q_i \in Q} CL_E(q_i, T)
    \label{eq:cle_definition}
\end{equation}
For each query $q_i$, its extraneous load $CL_E(q_i, T)$ is the sum of two normalized scores (each in $[0, 1]$) determined by Gemini-2.5-pro. These scores separately evaluate: 1) the query's ambiguity, and 2) the potential for distraction from irrelevant but plausible tools in the set $T$. A higher score in either component reflects greater cognitive load.
\subsection{Final Model and Accuracy Prediction}
By combining the intrinsic and extraneous components (Equation \eqref{eq:total_load_decomposition}), we arrive at the total cognitive load for a task instance $(Q, T, G)$. This unified metric allows us to predict model performance directly from our first postulate:
\begin{equation}
    \text{Accuracy}(Q, T, G) \approx \exp(-(k \cdot CL_{Total} + b))
    \label{eq:accuracy_prediction_final}
\end{equation}
This model provides a comprehensive, theoretically-grounded framework for quantifying task complexity and model capability in multi-turn tool agent systems.

\subsection{Dataset Construction}

To create a benchmark with fine-grained control over cognitive load, we constructed \textbf{ToolLoad-Bench}. Our methodology began with the 200 high-quality instances from the multi-turn base split of the Berkeley Function Calling Leaderboard (BFCL) v3 \cite{patil2025bfcl}. From this foundation, we first extracted the 
tool dependency relationships to form initial Tool Interaction Graphs.

\begin{table}[h]
\centering
\begin{tabular}{lccccc}
\toprule
\textbf{Benchmark} & \textbf{Num} & \textbf{Domains} & \textbf{Tools}  & \textbf{Avg. Calls} \\
\midrule
BFCL   & 200 & 8 & 84 &  4.1 \\
ToolLoad & 500 & 10 & 106 &  4.9 \\
\bottomrule
\end{tabular}%
\caption{Statistical comparison of ToolLoad-Bench and the original BFCL-v3 (multi-turn base)\cite{patil2025bfcl} dataset.}
\label{tab:dataset_comparison_transposed}
\end{table}

We then employed a novel graph-to-task generation pipeline to build a larger and more diverse dataset. This involved two key strategies: 1) \textbf{Graph Generation}, where we synthesized entirely new, complex task graphs, and 2) \textbf{Edge Insertion}, where we systematically added new dependency edges to existing graphs to increase their structural complexity. Furthermore, to address the lack of scenarios with highly complex dependencies, we designed two new tool categories and meticulously annotated new instances following the official BFCL-v3 protocol.

This pipeline resulted in a final dataset of 500 instances designed to push the limits of current models. The statistical profile of ToolLoad-Bench is summarized in Table \ref{tab:dataset_comparison_transposed}. The algorithms for data generation and cognitive load computation are detailed in the Appendix.

\section{Experiments and Analysis}

We conducted a series of experiments on our ToolLoad-Bench to evaluate leading language models and validate our cognitive load framework.

\subsection{Experimental Setup}
\paragraph{Models.} We evaluated a comprehensive suite of models to cover different capability tiers \cite{patil2025bfcl, chen2025acebench, chen2023t}.
\begin{itemize}
\item \textbf{Closed-Source models:} GPT-4o, GPT-4o-mini \cite{achiam2023gpt}, Gemini 2.5 Pro \cite{team2024gemini}, and Claude 3.7 Sonnet.
\item \textbf{Open-Source models:} A range of models from the Qwen3 \cite{yang2025qwen3} and Llama3.3 families \cite{dubey2024llama}, including Qwen3-8B, Qwen3-32B, Qwen3-235B, and Llama3.3-70B.
\item \textbf{Fine-tuned model} xLAM2-32B \cite{prabhakar2025apigen, zheng2024agentstudio}, a model specifically fine-tuned for advanced multi-turn tool use, to compare against general-purpose models.
\end{itemize}

\paragraph{Metrics.} Our primary evaluation metric is \textbf{Accuracy}, which measures the rate of successful task completion. The detailed methodology for calculating accuracy is provided in the Appendix.

\begin{table}[h]
\centering

\begin{tabular}{lc}
\toprule
\textbf{Model} & \textbf{Overall Accuracy (\%)} \\
\midrule
\multicolumn{2}{l}{\textit{Closed-Source models}} \\
GPT-4o & 68 \\
Claude 3.7 Sonnet & 64.8 \\
Gemini 2.5 Pro & 60 \\
GPT-4o-mini & 62.2 \\
\midrule
\multicolumn{2}{l}{\textit{Open-Source models}} \\
Qwen3-235B & 58 \\
Llama3.3-70B & 17 \\
Qwen3-32B & 55.2 \\
Qwen3-8B & 38.6 \\
\midrule
\multicolumn{2}{l}{\textit{Fine-tuned model}} \\
xLAM2-32B & \textbf{78.8} \\

\bottomrule
\end{tabular}
\caption{Overall Accuracy (\%) on ToolLoad-Bench.}
\label{tab:overall_performance}
\end{table}

\subsection{Overall Performance}
Table \ref{tab:overall_performance} presents the overall accuracy for each model across the entire ToolLoad-Bench dataset. The results immediately underscore the challenging nature of our benchmark. A clear performance hierarchy emerges: the leading closed-source models form a top tier, with GPT-4o at 68.0\%. Among the open-source models, performance generally scales with size, with Qwen3-235B outperforming its smaller variants.

However, the most striking result comes from the specialized fine-tuned model, xLAM2-32B, which achieves the highest accuracy at 78.8\%. Despite its smaller size, its focused training allows it to significantly outperform larger general-purpose models. This finding strongly suggests that targeted fine-tuning is a highly effective strategy for boosting tool-use capabilities. The wide performance delta across all models demonstrates that ToolLoad-Bench effectively differentiates model capabilities. 

\subsection{Impact of Cognitive Load}
To understand how different facets of complexity affect performance, we analyzed model accuracy as a function of both intrinsic and extraneous cognitive load. We partitioned the dataset into low, medium, and high load buckets for each load type, with each bucket containing one-third of the instances.

\begin{figure}[t]
\centering
\includegraphics[width=0.95\columnwidth]{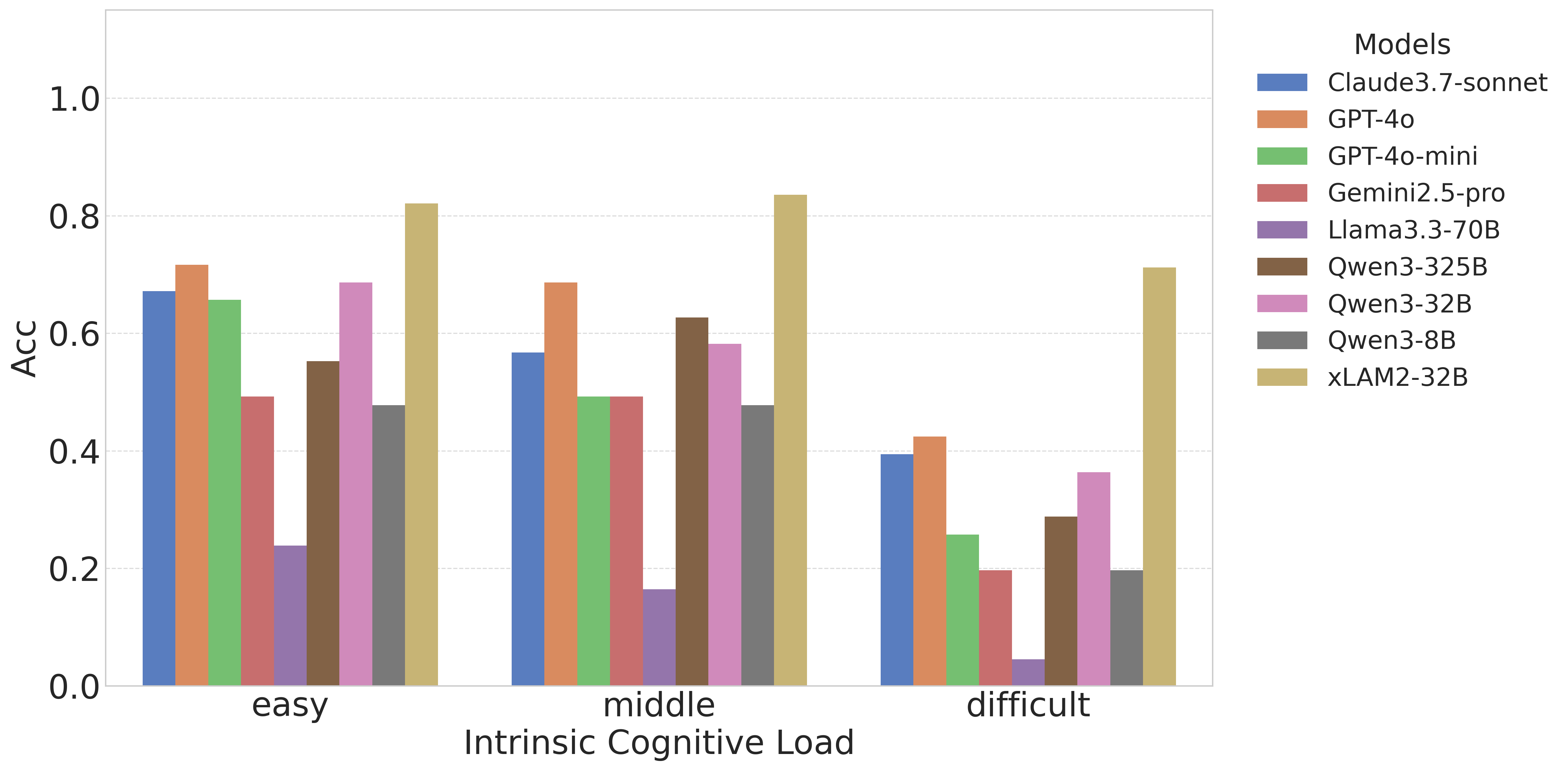} 
\caption{Accuracy vs. Intrinsic Cognitive Load ($CL_I$).}
\label{fig:cli_impact}
\end{figure}

\begin{figure}[t]
\centering
\includegraphics[width=0.95\columnwidth]{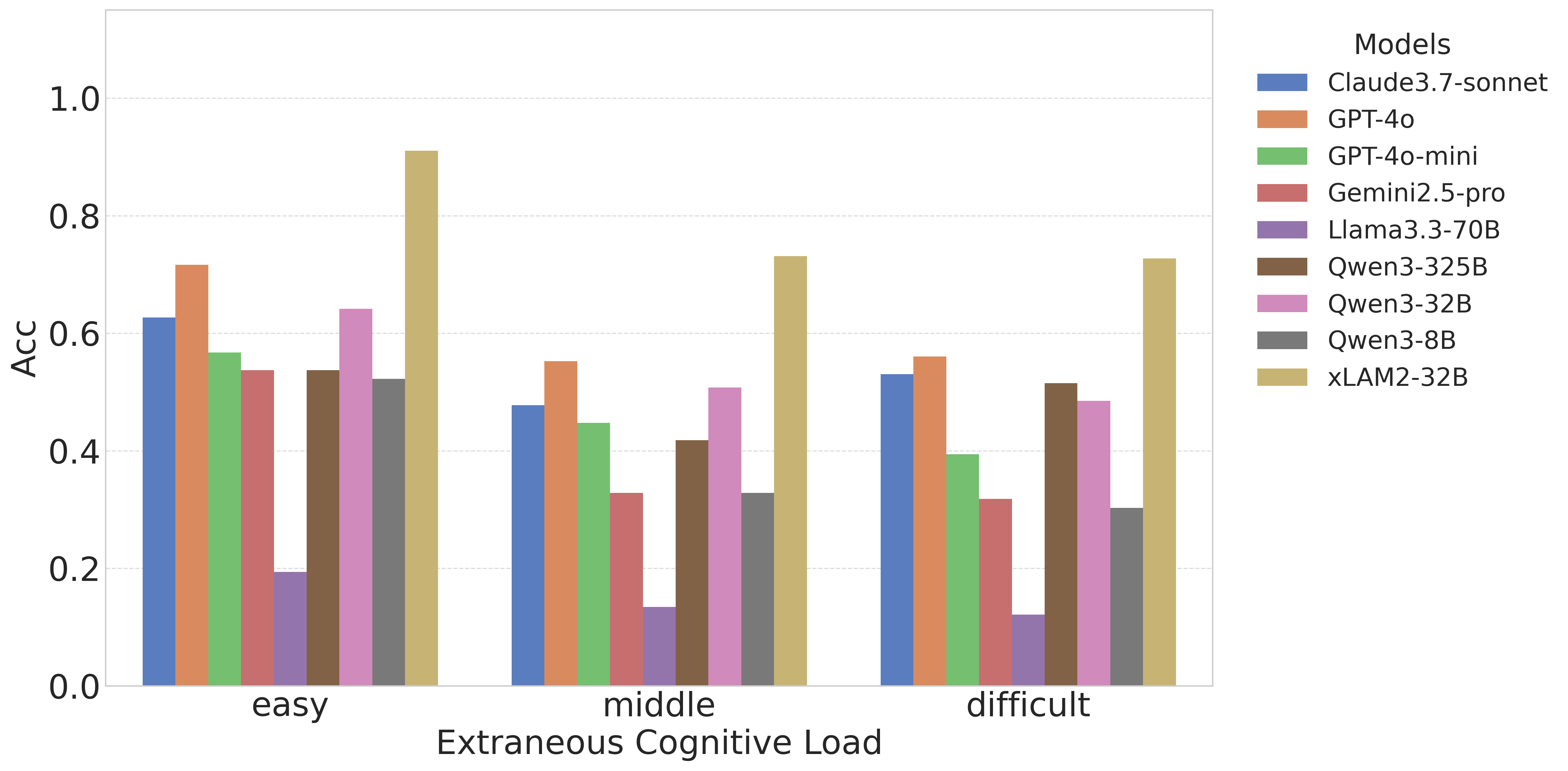}
\caption{Accuracy vs. Extraneous Cognitive Load ($CL_E$).}
\label{fig:cle_impact}
\end{figure}

\paragraph{Intrinsic Load Analysis.}
Figure \ref{fig:cli_impact} shows that accuracy consistently drops as the task's structural complexity ($CL_I$) increases. In the low-load regime, most models perform well, with the notable exception of Llama3.3-70B (23\% accuracy), indicating a fundamental weakness. At high loads, performance collapses for general-purpose models. Only the specialized xLAM2-32B maintains over 60\% accuracy, demonstrating its superior capability in handling complex reasoning structures.

\paragraph{Extraneous Load Analysis.}
A similar performance degradation is observed with rising Extraneous Cognitive Load ($CL_E$), as shown in Figure \ref{fig:cle_impact}. Higher query ambiguity and the presence of distractor tools consistently reduce accuracy across all models. The performance patterns mirror the $CL_I$ results: Llama3.3-70B again struggles, while only the fine-tuned xLAM2-32B sustains high accuracy under high extraneous load. This confirms that confusing task presentation is as significant a hurdle as inherent task complexity.

\subsection{Synthesizing Total Cognitive Load}

\begin{figure}[t]
\centering
\includegraphics[width=0.95\columnwidth]{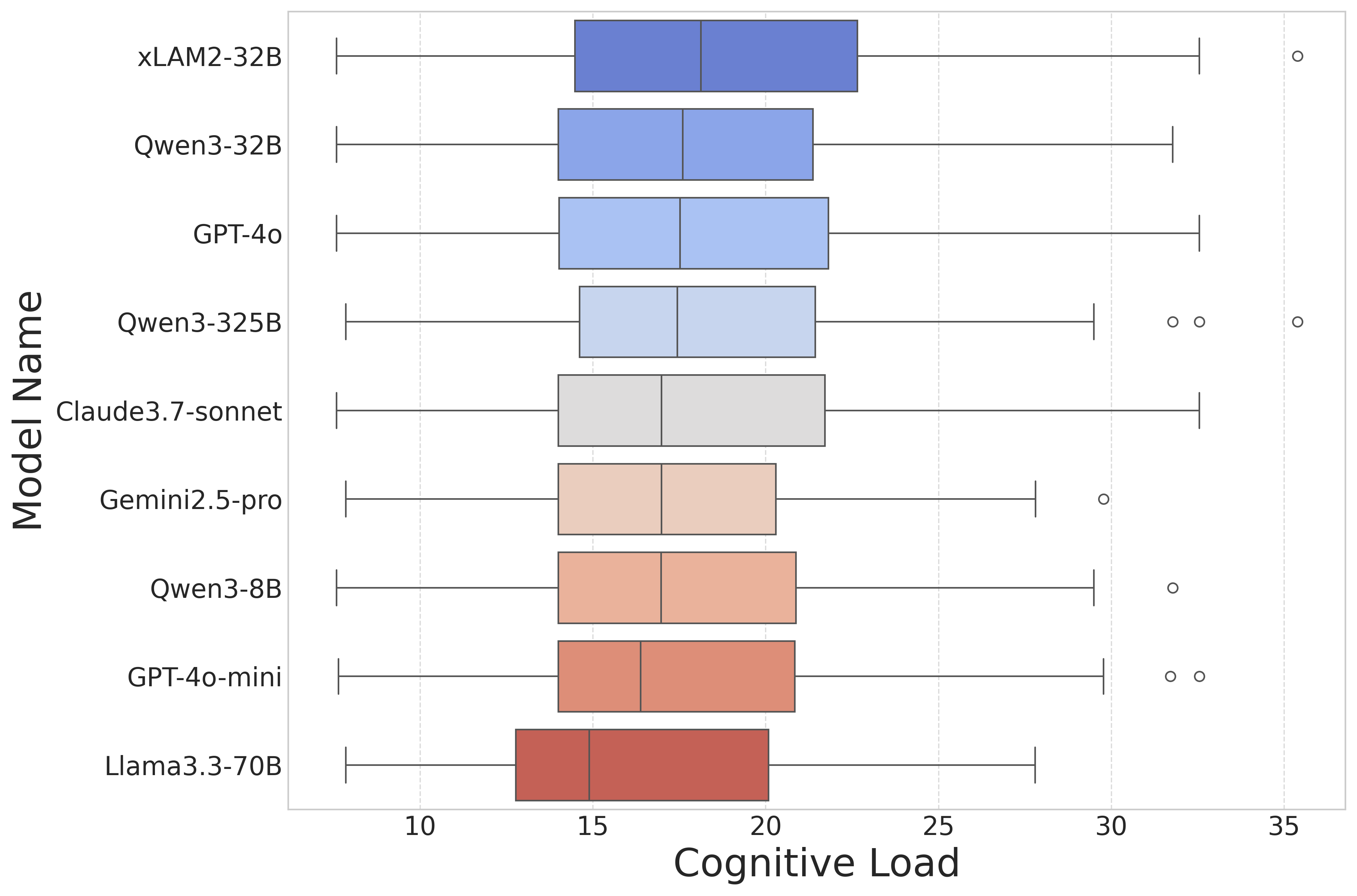}
\caption{Boxplots showing the distribution of Total Cognitive Load ($CL_{Total}$) for successfully completed tasks by different models.}
\label{fig:violin_load}
\end{figure}

% \begin{figure}[t]
% \centering
% \includegraphics[width=0.95\columnwidth]{LaTeX/figs/predictive.png}
% \caption{Predictive Accuracy vs cognitive load}
% \label{fig:predictive_accuracy}
% \end{figure}

\begin{figure*}[t] 
\centering % 
\includegraphics[width=\textwidth]{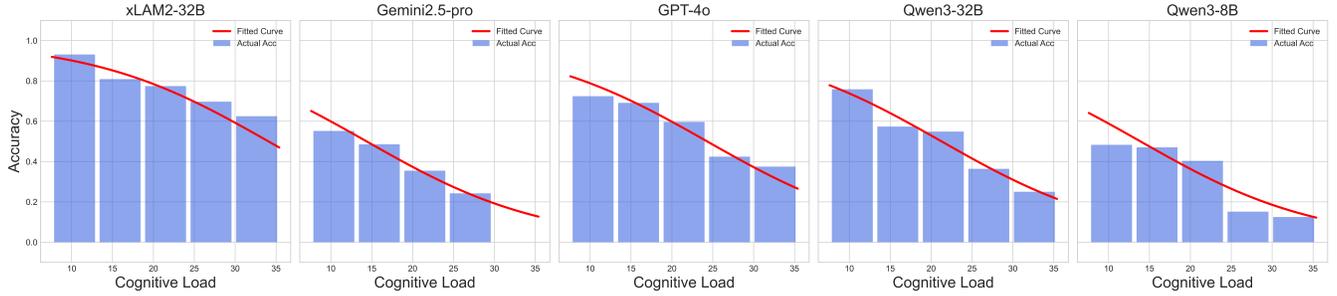} % 
\caption{Empirical accuracy vs. Total Cognitive Load. The blue bars show the actual accuracy within binned load intervals, while the red line represents the fitted exponential decay curve from our theoretical model.}\label{fig:load_accuracy_fit}
\end{figure*}

While analyzing intrinsic and extraneous loads separately provides valuable insights, a unified \textbf{Total Cognitive Load ($CL_{Total}$)} is required to holistically measure task difficulty and map a model's operational limits. The synthesis of these two components is not arbitrary but is directly guided by our Load-Success Relationship postulate (Equation \ref{eq:load_success_relation}).

The postulate implies that any two distinct cognitive challenges that cause an equivalent drop in success probability must correspond to an equivalent amount of cognitive load. This principle provides a direct method for calibrating the relative contributions of $CL_I$ and $CL_E$ onto a single, unified scale. We can empirically determine a scaling factor, $\omega_E$, that quantifies how much a unit of our measured extraneous load impacts accuracy relative to a unit of intrinsic load. This factor is calculated as the ratio of their observed effects on model performance:
\begin{equation}
    \omega_E = \frac{\Delta_{\text{Acc}}(CL_E)}{\Delta_{\text{Acc}}(CL_I)}
\end{equation}
where $\Delta_{\text{Acc}}(CL)$ represents the empirically measured drop in accuracy associated with an increase in that type of load. By setting the weight of intrinsic load to 1 as our baseline, this ratio places extraneous load on the same effective scale. The final, model-specific Total Cognitive Load is then a principled weighted sum:
\begin{equation}
    CL_{Total} = CL_I + \omega_E \cdot CL_E
\end{equation}
This unified score reflects how a specific model weighs the challenges of inherent task complexity versus ambiguous presentation, providing a single, powerful metric for defining its capability boundary.

\subsection{Defining Model Capability Limits with Total Cognitive Load}
Having established a unified $CL_{Total}$ score, we can move beyond overall accuracy to precisely map the operational boundaries of each model. This analysis proceeds in three steps: first, visualizing the range of solvable tasks; second, modeling the performance decay as a function of load; and third, extracting key parameters that define each model's cognitive profile.

\paragraph{Visualizing the Operational Range.}
We begin by visualizing the distribution of $CL_{Total}$ for only those tasks that each model \textit{successfully} completed. The boxplots in Figure \ref{fig:violin_load} powerfully illustrate the effective cognitive capacity of each model. The distributions reveal stark differences: less capable models like Llama3.3-70B are confined to a narrow band of low-load tasks. In contrast, top-performing agents like the specialized xLAM2-32B and GPT-4o exhibit distributions with a higher median and a significantly wider range. The upper quartile and whisker of each distribution serve as a clear visual signature of a model's capability.

\paragraph{Modeling the Performance Decay Curve.}
While the boxplots show the range of solvable problems, they don't capture the probability of accuracy at different cognitive loads. We fit our theoretical Load-Success Relationship (Equation \ref{eq:accuracy_prediction_final}) to the empirical data for each model. Figure \ref{fig:load_accuracy_fit} plots the actual accuracy (blue bars) across binned cognitive load levels against the fitted exponential decay curve (red line). The close alignment between the empirical data and the theoretical curve provides strong visual validation for our framework.

\paragraph{Quantifying Capability with Model-Specific Parameters.}
The fitted curves can be characterized by the parameters $k$ and $b$ from our core equation, $\text{Accuracy} \approx \exp(-(k \cdot CL_{Total} + b))$. These parameters provide a concise, quantitative summary of a model's cognitive profile:
\begin{itemize}
    \item \textbf{Baseline Capability ($b$):} This parameter reflects the model's intrinsic capability at near-zero cognitive load. A lower $b$ corresponds to a higher starting accuracy ($\text{Acc} \approx e^{-b}$ at $CL_{Total}=0$), indicating a stronger foundational ability.
    \item \textbf{Load Sensitivity ($k$):} This parameter measures how resilient the model is to increasing cognitive load. A smaller $k$ signifies a flatter decay curve, meaning the model's performance degrades more gracefully under pressure.
\end{itemize}

Table \ref{tab:model_parameters} presents the fitted $k$ and $b$ values for key models. The specialized xLAM2-32B exhibits the lowest $k$ and a very low $b$, quantifying its dual strength: the highest baseline proficiency and exceptional robustness to complexity. In contrast, GPT-4o shows a similarly low sensitivity but a slightly higher baseline load, suggesting it is highly capable on simpler tasks but its performance degrades more quickly than the specialized fine-tuned model. The open source Qwen3 models show a clear progression, with the 235B model approaching the capability of closed-source giants, while the smaller 8B model has both a weaker baseline and higher sensitivity. This parametric analysis transforms the abstract notion of "capability" into a concrete, two-dimensional profile, precisely defining each agent's strengths and breaking points.

\begin{table}[h]
\centering
\begin{tabular}{lcc}
\toprule
\textbf{Model} & Load Sensitivity($k$) & Baseline Load($b$) \\
\midrule
xLAM2-32B & \textbf{0.034} & 1.22 \\
GPT-4o & 0.067 & 1.71 \\
Claude 3.7 & 0.073 & 1.57 \\
Gemini2.5-pro & 0.088 & 1.22 \\
Qwen3-32B & 0.075 & 1.60 \\
Qwen3-8B & 0.085 & \textbf{1.12} \\
\bottomrule
\end{tabular}
\caption{Fitted cognitive load parameters for different models. Lower $k$ indicates better resilience to load, and lower $b$ indicates higher baseline accuracy.}
\label{tab:model_parameters}
\end{table}

\subsection{Validating the Cognitive Load Distributional Model}
\label{sec:distribution_validation}

Our framework's central hypothesis is that cognitive load shapes the \textit{probability distribution} of accuracy, not that it deterministically predicts an outcome. To validate this, we assess the \textbf{goodness-of-fit} between our model's predicted probabilities and the empirically observed accuracy. We use two complementary methods for this validation.

First, we employ the formal Hosmer-Lemeshow (H-L) statistical test \cite{paul2013standardizing}. For this test, the null hypothesis is that the model is well-calibrated, meaning a high p-value is the desired outcome. As shown in Table \ref{tab:hosmer_lemeshow}, the p-values for all evaluated models are well above the conventional 0.05 significance level. This provides strong statistical evidence that our framework generates a probability distribution of accuracy that is statistically indistinguishable from the observed reality.

\begin{table}[ht]
\centering
\begin{tabular}{lcc}
\toprule
\textbf{Model} & \textbf{H-L $\chi^2$ Statistic} & \textbf{p-value} \\
\midrule
\multicolumn{3}{l}{\textit{Closed-Source models}} \\
GPT-4o & 4.87 & 0.77 \\
Claude 3.7 Sonnet & 10.47 & 0.23 \\
Gemini 2.5 Pro & 13.15 & 0.11 \\
GPT-4o-mini & 8.91 & 0.35 \\
\midrule
\multicolumn{3}{l}{\textit{Open-Source models}} \\
Qwen3-235B & 5.19 & 0.74 \\
Llama3.3-70B & 13.21 & 0.10 \\
Qwen3-32B & 7.50 & 0.48 \\
Qwen3-8B & 7.90 & 0.44 \\
\midrule
\multicolumn{3}{l}{\textit{Finetuned model}} \\
xLAM2-32B & \textbf{3.59} & \textbf{0.89} \\
\bottomrule
\end{tabular}
\caption{Hosmer-Lemeshow Goodness-of-Fit Test Results.}
\label{tab:hosmer_lemeshow}
\end{table}

\begin{figure}[t]
\centering
\includegraphics[width=0.95\columnwidth]{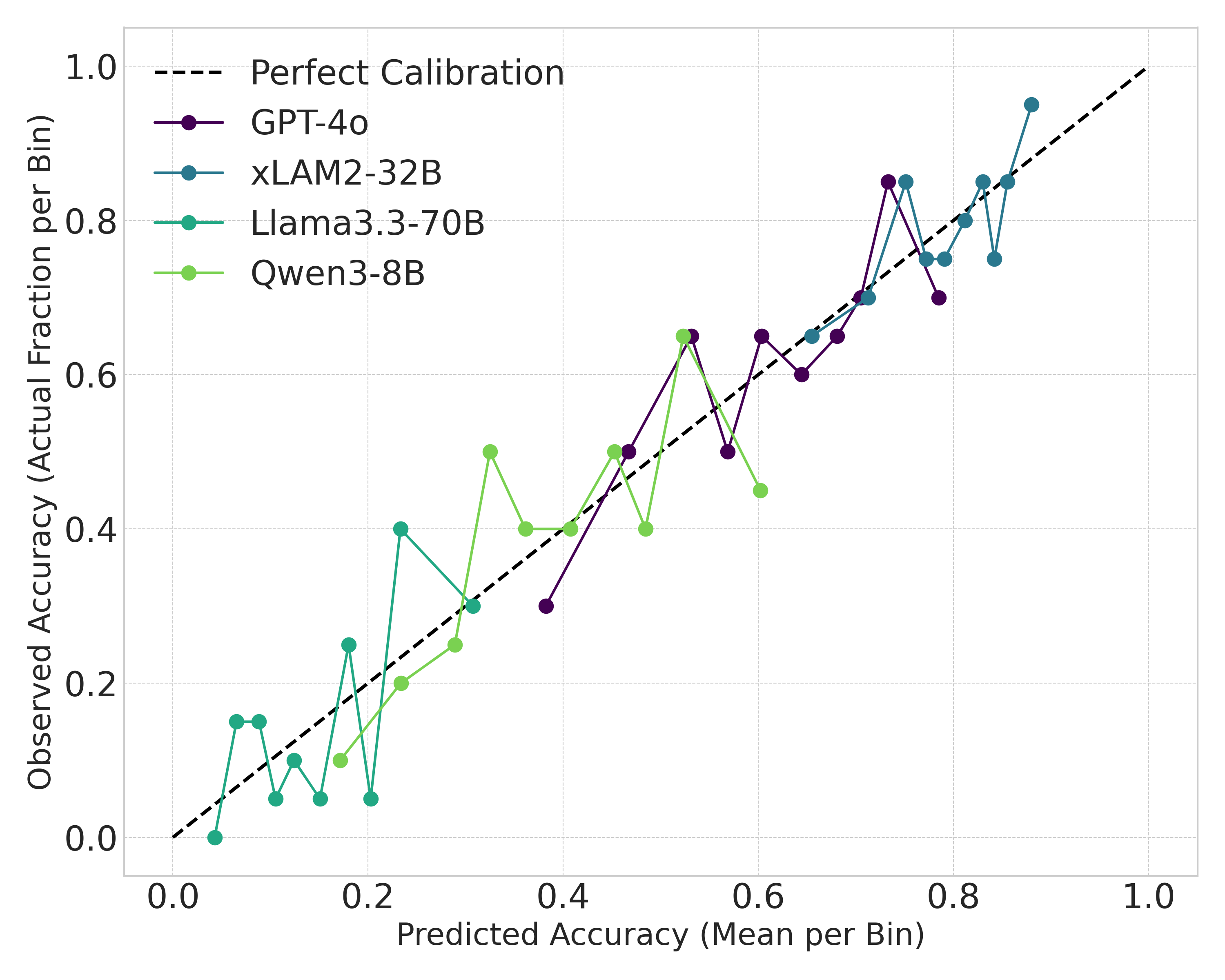}
\caption{Calibration plot for representative models, showing predicted probabilities vs. observed accuracy.}
\label{fig:calibration_plot}
\end{figure}

Second, we visually corroborate this statistical finding with calibration plots, as shown in Figure \ref{fig:calibration_plot}. The plots reveal a close alignment between the predicted probabilities and the observed accuracy, with points lying near the diagonal line of perfect calibration. Together, these results validate our framework's foundational assumption, demonstrating that it accurately models the nuanced, probabilistic nature of tool-use accuracy.

\section{Discussion}

\paragraph{Theoretical Contribution: A Scientific Framework for Evaluation.}
From a theoretical standpoint, our primary contribution is the introduction and validation of cognitive load as a formal construct for measuring task complexity in tool-use scenarios. We moved beyond abstract notions of difficulty by operationalizing it into measurable components: Intrinsic Load ($CL_I$) from the task's inherent structure and Extraneous Load ($CL_E$) from its presentation. To test this theory, we constructed \textbf{ToolLoad-Bench}, a benchmark specifically designed for controlled experimentation. Our experiments (Figures \ref{fig:cli_impact} and \ref{fig:cle_impact}) provide strong empirical evidence for our central hypothesis: a direct and predictable relationship exists between a task's cognitive load and a model's accuracy. This provides the research community with a more scientific and comprehensive methodology for evaluation, enabling a shift from simple leaderboards to diagnostic analysis.

\paragraph{Application Value: Enabling Intelligent Task Routing.}
From a practical standpoint, the ability to quantify cognitive load offers immediate application value. Our framework quantitatively reveals tool-use agents' distinct capabilities for handling tasks with varying cognitive loads. This provides a principled basis for a critical real-world application: intelligent task routing \cite{yue2025masrouter, hu2024routerbench}. In a production system, it could dynamically route the task to the most appropriate LLM based on cognitive scores. This approach enables the design of highly efficient, scalable, and economically viable tool-use agent systems.

\paragraph{Limitations and Future Work.}
Finally, we acknowledge the limitations of our current framework. While ToolLoad-Bench is highly controlled, its domain coverage could be expanded to ensure broader generalizability. Furthermore, our measurement of extraneous load ($CL_E$) currently relies on LLM evaluation, and developing more objective, feature-based metrics would strengthen the framework.

\section{Conclusion}

This work challenges the prevailing evaluation paradigm for tool-augmented LLMs, which reduces agent capability to a single, opaque score. We introduce a diagnostic framework grounded in Cognitive Load Theory that moves beyond simple accuracy to map the capability boundaries of tool-use agents. Our central finding is that models possess distinct cognitive frontiers, exhibiting sharp, predictable performance cliffs as task complexity increases. Compared to traditional tool-use agent evaluations, our research provides a principled methodology for understanding their true limits. 

\section{Acknowledgements}
 This work was supported by the National Natural Science Foundation of China(No.U21B2009).

% I will skip acknowledgements for an anonymous submission.
\iffalse 
\section*{Acknowledgments}
This work was supported by... We would like to thank...
\fi

\bibliography{aaai2026}

@misc{api_bank,
      title={API-Bank: A Comprehensive Benchmark for Tool-Augmented LLMs}, 
      author={Minghao Li and Yingxiu Zhao and Bowen Yu and Feifan Song and Hangyu Li and Haiyang Yu and Zhoujun Li and Fei Huang and Yongbin Li},
      year={2023},
      eprint={2304.08244},
      archivePrefix={arXiv},
      primaryClass={cs.CL},
      url={https://arxiv.org/abs/2304.08244}, 
}

@misc{gorilla,
      title={Gorilla: Large Language Model Connected with Massive APIs}, 
      author={Shishir G. Patil and Tianjun Zhang and Xin Wang and Joseph E. Gonzalez},
      year={2023},
      eprint={2305.15334},
      archivePrefix={arXiv},
      primaryClass={cs.CL},
      url={https://arxiv.org/abs/2305.15334}, 
}

@inproceedings{hugginggpt,
  author = {Shen, Yongliang and Song, Kaitao and Tan, Xu and Li, Dongsheng and Lu, Weiming and Zhuang, Yueting},
  booktitle = {Advances in Neural Information Processing Systems},
  title = {HuggingGPT: Solving AI Tasks with ChatGPT and its Friends in HuggingFace},
  year = {2023}
}

@article{lin2024hammer,
 author = {Lin, Qiqiang and Wen, Muning and Peng, Qiuying and Nie, Guanyu and Liao, Junwei and Wang, Jun and Mo, Xiaoyun and Zhou, Jiamu and Cheng, Cheng and Zhao, Yin and others},
 journal = {ArXiv preprint},
 title = {Hammer: Robust function-calling for on-device language models via function masking},
 url = {https://arxiv.org/abs/2410.04587},
 volume = {abs/2410.04587},
 year = {2024}
}

@inproceedings{schick2023toolformer,
 author = {Timo Schick and
Jane Dwivedi{-}Yu and
Roberto Dess{\`{\i}} and
Roberta Raileanu and
Maria Lomeli and
Eric Hambro and
Luke Zettlemoyer and
Nicola Cancedda and
Thomas Scialom},
 bibsource = {dblp computer science bibliography, https://dblp.org},
 booktitle = {Advances in Neural Information Processing Systems 36: Annual Conference
on Neural Information Processing Systems 2023, NeurIPS 2023, New Orleans,
LA, USA, December 10 - 16, 2023},
 editor = {Alice Oh and
Tristan Naumann and
Amir Globerson and
Kate Saenko and
Moritz Hardt and
Sergey Levine},
 title = {Toolformer: Language Models Can Teach Themselves to Use Tools},
 url = {http://papers.nips.cc/paper\_files/paper/2023/hash/d842425e4bf79ba039352da0f658a906-Abstract-Conference.html},
 year = {2023}
}

@article{sweller1988cognitive,
  title={Cognitive load during problem solving: Effects on learning},
  author={Sweller, John},
  journal={Cognitive science},
  volume={12},
  number={2},
  pages={257--285},
  year={1988},
  publisher={Wiley Online Library}
}

@misc{toolalpaca,
      title={ToolAlpaca: Generalized Tool Learning for Language Models with 3000 Simulated Cases}, 
      author={Qiaoyu Tang and Ziliang Deng and Hongyu Lin and Xianpei Han and Qiao Liang and Boxi Cao and Le Sun},
      year={2023},
      eprint={2306.05301},
      archivePrefix={arXiv},
      primaryClass={cs.CL},
      url={https://arxiv.org/abs/2306.05301}, 
}

@misc{toolbench,
      title={ToolLLM: Facilitating Large Language Models to Master 16000+ Real-world APIs}, 
      author={Yujia Qin and Shihao Liang and Yining Ye and Kunlun Zhu and Lan Yan and Yaxi Lu and Yankai Lin and Xin Cong and Xiangru Tang and Bill Qian and Sihan Zhao and Lauren Hong and Runchu Tian and Ruobing Xie and Jie Zhou and Mark Gerstein and Dahai Li and Zhiyuan Liu and Maosong Sun},
      year={2023},
      eprint={2307.16789},
      archivePrefix={arXiv},
      primaryClass={cs.AI},
      url={https://arxiv.org/abs/2307.16789}, 
}

@misc{tooltalk,
      title={ToolTalk: Evaluating Tool-Usage in a Conversational Setting}, 
      author={Nicholas Farn and Richard Shin},
      year={2023},
      eprint={2311.10775},
      archivePrefix={arXiv},
      primaryClass={cs.CL},
      url={https://arxiv.org/abs/2311.10775}, 
}

@inproceedings{yaoreact,
 author = {Shunyu Yao and
Jeffrey Zhao and
Dian Yu and
Nan Du and
Izhak Shafran and
Karthik R. Narasimhan and
Yuan Cao},
 bibsource = {dblp computer science bibliography, https://dblp.org},
 booktitle = {The Eleventh International Conference on Learning Representations,
{ICLR} 2023, Kigali, Rwanda, May 1-5, 2023},
 publisher = {OpenReview.net},
 title = {ReAct: Synergizing Reasoning and Acting in Language Models},
 url = {https://openreview.net/pdf?id=WE\_vluYUL-X},
 year = {2023}
}

@article{plass2010cognitive,
  title={Cognitive load theory},
  author={Plass, Jan L and Moreno, Roxana and Br{\"u}nken, Roland},
  year={2010},
  publisher={Cambridge university press}
}

@article{prabhakar2025apigen,
  title={Apigen-mt: Agentic pipeline for multi-turn data generation via simulated agent-human interplay},
  author={Prabhakar, Akshara and Liu, Zuxin and Zhu, Ming and Zhang, Jianguo and Awalgaonkar, Tulika and Wang, Shiyu and Liu, Zhiwei and Chen, Haolin and Hoang, Thai and Niebles, Juan Carlos and others},
  journal={arXiv preprint arXiv:2504.03601},
  year={2025}
}

@inproceedings{patil2025bfcl,
title={The Berkeley Function Calling Leaderboard (BFCL): From Tool Use to Agentic Evaluation of Large Language Models}, 
author={Patil, Shishir G. and Mao, Huanzhi and Cheng-Jie Ji, Charlie and Yan, Fanjia and Suresh, Vishnu and Stoica, Ion and E. Gonzalez, Joseph},
booktitle={Forty-second International Conference on Machine Learning},
year={2025},
}

@article{yin2025magnet,
  title={Magnet: Multi-turn tool-use data synthesis and distillation via graph translation},
  author={Yin, Fan and Wang, Zifeng and Hsu, I and Yan, Jun and Jiang, Ke and Chen, Yanfei and Gu, Jindong and Le, Long T and Chang, Kai-Wei and Lee, Chen-Yu and others},
  journal={arXiv preprint arXiv:2503.07826},
  year={2025}
}

@article{qian2025toolrl,
  title={Toolrl: Reward is all tool learning needs},
  author={Qian, Cheng and Acikgoz, Emre Can and He, Qi and Wang, Hongru and Chen, Xiusi and Hakkani-T{\"u}r, Dilek and Tur, Gokhan and Ji, Heng},
  journal={arXiv preprint arXiv:2504.13958},
  year={2025}
}

@article{dong2025tool,
  title={Tool-Star: Empowering LLM-Brained Multi-Tool Reasoner via Reinforcement Learning},
  author={Dong, Guanting and Chen, Yifei and Li, Xiaoxi and Jin, Jiajie and Qian, Hongjin and Zhu, Yutao and Mao, Hangyu and Zhou, Guorui and Dou, Zhicheng and Wen, Ji-Rong},
  journal={arXiv preprint arXiv:2505.16410},
  year={2025}
}

@article{shi2025taskcraft,
  title={Taskcraft: Automated generation of agentic tasks},
  author={Shi, Dingfeng and Cao, Jingyi and Chen, Qianben and Sun, Weichen and Li, Weizhen and Lu, Hongxuan and Dong, Fangchen and Qin, Tianrui and Zhu, King and Liu, Minghao and others},
  journal={arXiv preprint arXiv:2506.10055},
  year={2025}
}

@article{dubey2024llama,
 author = {Dubey, Abhimanyu and Jauhri, Abhinav and Pandey, Abhinav and Kadian, Abhishek and Al-Dahle, Ahmad and Letman, Aiesha and Mathur, Akhil and Schelten, Alan and Yang, Amy and Fan, Angela and others},
 journal = {ArXiv preprint},
 title = {The llama 3 herd of models},
 url = {https://arxiv.org/abs/2407.21783},
 volume = {abs/2407.21783},
 year = {2024}
}

@article{yang2025qwen3,
  title={Qwen3 technical report},
  author={Yang, An and Li, Anfeng and Yang, Baosong and Zhang, Beichen and Hui, Binyuan and Zheng, Bo and Yu, Bowen and Gao, Chang and Huang, Chengen and Lv, Chenxu and others},
  journal={arXiv preprint arXiv:2505.09388},
  year={2025}
}

@article{zheng2024agentstudio,
  title={Agentstudio: A toolkit for building general virtual agents},
  author={Zheng, Longtao and Huang, Zhiyuan and Xue, Zhenghai and Wang, Xinrun and An, Bo and Yan, Shuicheng},
  journal={arXiv preprint arXiv:2403.17918},
  year={2024}
}

@article{achiam2023gpt,
 author = {Achiam, Josh and Adler, Steven and Agarwal, Sandhini and Ahmad, Lama and Akkaya, Ilge and Aleman, Florencia Leoni and Almeida, Diogo and Altenschmidt, Janko and Altman, Sam and Anadkat, Shyamal and others},
 journal = {ArXiv preprint},
 title = {Gpt-4 technical report},
 url = {https://arxiv.org/abs/2303.08774},
 volume = {abs/2303.08774},
 year = {2023}
}

@article{team2024gemini,
 author = {Team, Gemini and Georgiev, Petko and Lei, Ving Ian and Burnell, Ryan and Bai, Libin and Gulati, Anmol and Tanzer, Garrett and Vincent, Damien and Pan, Zhufeng and Wang, Shibo and others},
 journal = {ArXiv preprint},
 title = {Gemini 1.5: Unlocking multimodal understanding across millions of tokens of context},
 url = {https://arxiv.org/abs/2403.05530},
 volume = {abs/2403.05530},
 year = {2024}
}

@article{guidroz2025llm,
  title={LLM-based Text Simplification and its Effect on User Comprehension and Cognitive Load},
  author={Guidroz, Theo and Ardila, Diego and Li, Jimmy and Mansour, Adam and Jhun, Paul and Gonzalez, Nina and Ji, Xiang and Sanchez, Mike and Kakarmath, Sujay and Bellaiche, Mathias MJ and others},
  journal={arXiv preprint arXiv:2505.01980},
  year={2025}
}

@article{xiao2025streaming,
  title={Streaming, Fast and Slow: Cognitive Load-Aware Streaming for Efficient LLM Serving},
  author={Xiao, Chang and Yang, Brenda},
  journal={arXiv preprint arXiv:2504.17999},
  year={2025}
}

@article{sirbu2025explanation,
  title={Explanation Provision Strategies in LLM-based Data Assistants: Impact on Extraneous Cognitive Load, Trust, and Task Performance},
  author={S{\^\i}rbu, Ana-Maria and Schelhorn, Till Carlo and Gnewuch, Ulrich},
  year={2025}
}

@article{yang2025sparse,
  title={Sparse Brains are Also Adaptive Brains: Cognitive-Load-Aware Dynamic Activation for LLMs},
  author={Yang, Yiheng and Wang, Yujie and Ma, Chi and Yu, Lei and Chersoni, Emmanuele and Huang, Chu-Ren},
  journal={arXiv preprint arXiv:2502.19078},
  year={2025}
}

@article{paul2013standardizing,
  title={Standardizing the power of the Hosmer--Lemeshow goodness of fit test in large data sets},
  author={Paul, Prabasaj and Pennell, Michael L and Lemeshow, Stanley},
  journal={Statistics in medicine},
  volume={32},
  number={1},
  pages={67--80},
  year={2013},
  publisher={Wiley Online Library}
}

@article{yue2025masrouter,
  title={Masrouter: Learning to route llms for multi-agent systems},
  author={Yue, Yanwei and Zhang, Guibin and Liu, Boyang and Wan, Guancheng and Wang, Kun and Cheng, Dawei and Qi, Yiyan},
  journal={arXiv preprint arXiv:2502.11133},
  year={2025}
}

@article{hu2024routerbench,
  title={Routerbench: A benchmark for multi-llm routing system},
  author={Hu, Qitian Jason and Bieker, Jacob and Li, Xiuyu and Jiang, Nan and Keigwin, Benjamin and Ranganath, Gaurav and Keutzer, Kurt and Upadhyay, Shriyash Kaustubh},
  journal={arXiv preprint arXiv:2403.12031},
  year={2024}
}

@article{dong2025agentic,
  title={Agentic Reinforced Policy Optimization},
  author={Dong, Guanting and Mao, Hangyu and Ma, Kai and Bao, Licheng and Chen, Yifei and Wang, Zhongyuan and Chen, Zhongxia and Du, Jiazhen and Wang, Huiyang and Zhang, Fuzheng and others},
  journal={arXiv preprint arXiv:2507.19849},
  year={2025}
}

@article{du2024anytool,
  title={Anytool: Self-reflective, hierarchical agents for large-scale api calls},
  author={Du, Yu and Wei, Fangyun and Zhang, Hongyang},
  journal={arXiv preprint arXiv:2402.04253},
  year={2024}
}

@article{qu2025tool,
  title={Tool learning with large language models: A survey},
  author={Qu, Changle and Dai, Sunhao and Wei, Xiaochi and Cai, Hengyi and Wang, Shuaiqiang and Yin, Dawei and Xu, Jun and Wen, Ji-Rong},
  journal={Frontiers of Computer Science},
  volume={19},
  number={8},
  pages={198343},
  year={2025},
  publisher={Springer}
}

@article{mialon2023augmented,
  title={Augmented language models: a survey},
  author={Mialon, Gr{\'e}goire and Dess{\`\i}, Roberto and Lomeli, Maria and Nalmpantis, Christoforos and Pasunuru, Ram and Raileanu, Roberta and Rozi{\`e}re, Baptiste and Schick, Timo and Dwivedi-Yu, Jane and Celikyilmaz, Asli and others},
  journal={arXiv preprint arXiv:2302.07842},
  year={2023}
}

@article{wang2024survey,
  title={A survey on large language model based autonomous agents},
  author={Wang, Lei and Ma, Chen and Feng, Xueyang and Zhang, Zeyu and Yang, Hao and Zhang, Jingsen and Chen, Zhiyuan and Tang, Jiakai and Chen, Xu and Lin, Yankai and others},
  journal={Frontiers of Computer Science},
  volume={18},
  number={6},
  pages={186345},
  year={2024},
  publisher={Springer}
}

@article{qin2024tool,
  title={Tool learning with foundation models},
  author={Qin, Yujia and Hu, Shengding and Lin, Yankai and Chen, Weize and Ding, Ning and Cui, Ganqu and Zeng, Zheni and Zhou, Xuanhe and Huang, Yufei and Xiao, Chaojun and others},
  journal={ACM Computing Surveys},
  volume={57},
  number={4},
  pages={1--40},
  year={2024},
  publisher={ACM New York, NY}
}

@article{zhang2025nemotron,
  title={Nemotron-research-tool-n1: Tool-using language models with reinforced reasoning},
  author={Zhang, Shaokun and Dong, Yi and Zhang, Jieyu and Kautz, Jan and Catanzaro, Bryan and Tao, Andrew and Wu, Qingyun and Yu, Zhiding and Liu, Guilin},
  journal={arXiv preprint arXiv:2505.00024},
  year={2025}
}

@article{wang2025otc,
  title={Otc: Optimal tool calls via reinforcement learning},
  author={Wang, Hongru and Qian, Cheng and Zhong, Wanjun and Chen, Xiusi and Qiu, Jiahao and Huang, Shijue and Jin, Bowen and Wang, Mengdi and Wong, Kam-Fai and Ji, Heng},
  journal={arXiv e-prints},
  pages={arXiv--2504},
  year={2025}
}

@article{feng2025retool,
  title={Retool: Reinforcement learning for strategic tool use in llms},
  author={Feng, Jiazhan and Huang, Shijue and Qu, Xingwei and Zhang, Ge and Qin, Yujia and Zhong, Baoquan and Jiang, Chengquan and Chi, Jinxin and Zhong, Wanjun},
  journal={arXiv preprint arXiv:2504.11536},
  year={2025}
}

@article{su2025openthinkimg,
  title={Openthinkimg: Learning to think with images via visual tool reinforcement learning},
  author={Su, Zhaochen and Li, Linjie and Song, Mingyang and Hao, Yunzhuo and Yang, Zhengyuan and Zhang, Jun and Chen, Guanjie and Gu, Jiawei and Li, Juntao and Qu, Xiaoye and others},
  journal={arXiv preprint arXiv:2505.08617},
  year={2025}
}

@article{shen2024small,
  title={Small llms are weak tool learners: A multi-llm agent},
  author={Shen, Weizhou and Li, Chenliang and Chen, Hongzhan and Yan, Ming and Quan, Xiaojun and Chen, Hehong and Zhang, Ji and Huang, Fei},
  journal={arXiv preprint arXiv:2401.07324},
  year={2024}
}

@inproceedings{shi2025tool,
  title={Tool learning in the wild: Empowering language models as automatic tool agents},
  author={Shi, Zhengliang and Gao, Shen and Yan, Lingyong and Feng, Yue and Chen, Xiuyi and Chen, Zhumin and Yin, Dawei and Verberne, Suzan and Ren, Zhaochun},
  booktitle={Proceedings of the ACM on Web Conference 2025},
  pages={2222--2237},
  year={2025}
}

@article{shi2024learning,
  title={Learning to use tools via cooperative and interactive agents},
  author={Shi, Zhengliang and Gao, Shen and Chen, Xiuyi and Feng, Yue and Yan, Lingyong and Shi, Haibo and Yin, Dawei and Ren, Pengjie and Verberne, Suzan and Ren, Zhaochun},
  journal={arXiv preprint arXiv:2403.03031},
  year={2024}
}

@article{chen2025tool,
  title={Tool-as-interface: Learning robot policies from human tool usage through imitation learning},
  author={Chen, Haonan and Zhu, Cheng and Li, Yunzhu and Driggs-Campbell, Katherine},
  journal={arXiv preprint arXiv:2504.04612},
  year={2025}
}

@article{chen2025acebench,
  title={ACEBench: Who Wins the Match Point in Tool Learning?},
  author={Chen, Chen and Hao, Xinlong and Liu, Weiwen and Huang, Xu and Zeng, Xingshan and Yu, Shuai and Li, Dexun and Wang, Shuai and Gan, Weinan and Huang, Yuefeng and others},
  journal={arXiv e-prints},
  pages={arXiv--2501},
  year={2025}
}

@article{chen2023t,
  title={T-eval: Evaluating the tool utilization capability of large language models step by step},
  author={Chen, Zehui and Du, Weihua and Zhang, Wenwei and Liu, Kuikun and Liu, Jiangning and Zheng, Miao and Zhuo, Jingming and Zhang, Songyang and Lin, Dahua and Chen, Kai and others},
  journal={arXiv preprint arXiv:2312.14033},
  year={2023}
}

@article{chen2025toleap,
  title={ToLeaP: Rethinking Development of Tool Learning with Large Language Models},
  author={Chen, Haotian and Song, Zijun and Niu, Boye and Zhang, Ke and Ou, Litu and Lu, Yaxi and Zhang, Zhong and Cong, Xin and Lin, Yankai and Liu, Zhiyuan and others},
  journal={arXiv preprint arXiv:2505.11833},
  year={2025}
}

\end{document}